\documentclass[journal]{IEEEtran}

\ifCLASSINFOpdf

\else

\fi

\usepackage{array}
\usepackage{amsmath}
\usepackage{graphicx}
\usepackage{upgreek}
\usepackage{multirow}
\usepackage{color}
\usepackage{graphicx}
\usepackage{upgreek}
\usepackage{multirow}
\usepackage{bbding}

\begin{document}

\title{RGB-D Based Action Recognition with Light-weight 3D Convolutional Networks}

\author{Haokui~Zhang,~Ying~Li,~Peng~Wang, Yu Liu,~and~Chunhua~Shen%
\thanks{This work was supported by the National Natural Science Foundation of China (Grant No. 61871460, 6187023557), the National Key Research and Development Program of China (Grant No. 2016YFB0502502), The Foundation Project for Advanced Research Field of China (Grant No. 614023804016HK03002), and the Innovation Foundation for Doctor Dissertation of Northwestern Polytechnical University (Grant No. CX201816).}
\thanks{H.~Zhang, Y.~Li and P.~Wang are with the Shaanxi Provincial Key Lab of Speech and Image Information Processing, School of Computer Science, Northwestern Polytechnical University, Xi'an 710129, China (e-mail: hkzhang1991@mail.nwpu.edu.cn; lybyp@nwpu.edu.cn; peng.wang@nwpu.edu.cn;). This work was done when H. Zhang was visiting The University of Adelaide. (Corresponding author: Peng Wang)}%
\thanks{Y.~Liu and C.~Shen are with the School of Computer Science, the University of Adelaide, Adelaide, SA 5005, Australia (yu.liu04@adelaide.edu.au; e-mail:chunhua.shen@adelaide.edu.au).}%
}

\markboth{Manuscript}%
{Nov.\ 2018}

\maketitle

\begin{abstract}

Different from RGB videos, depth data in RGB-D videos provide key complementary information for tristimulus visual data which potentially could achieve accuracy improvement for action recognition. However, most of the existing action recognition models solely using RGB videos limit the performance capacity. Additionally, the state-of-the-art action recognition models, namely 3D convolutional neural networks (3D-CNNs) contain tremendous parameters suffering from computational inefficiency. In this paper, we propose a series of 3D light-weight architectures for action recognition based on RGB-D data. Compared with conventional 3D-CNN models, the proposed light-weight 3D-CNNs have considerably less parameters involving lower computation cost, while it results in favorable recognition performance. Experimental results on two public benchmark datasets show that our models can approximate or outperform the state-of-the-art approaches. Specifically, on the RGB+D-NTU (NTU) dataset, we achieve 93.2\% and 97.6\% for cross-subject and cross-view measurement, and on the Northwestern-UCLA Multiview Action 3D (N-UCLA) dataset, we achieve 95.5\% accuracy of cross-view.

\end{abstract}


\IEEEpeerreviewmaketitle

\tableofcontents
\clearpage

\section{Introduction}
\label{introduction}

\IEEEPARstart{A}{ction} recognition is one of the most active research fields in the computer vision community, and the broad applications include robotics \cite{yu2013vision}, video
 surveillance \cite{ahmed2015improved},  and medical caring \cite{marks2011system}.

For action recognition, in terms of feature extraction and feature representation, the two categories of approaches for action recognition are based on conventional handcrafted
 features and deep learning features, respectively. Both the conventional and deep learning based approaches have achieved remarkable improvements for action recognition  recently.
 A migration from the handcrafted features based approaches to deep learning based methods happens since the emerge of AlexNet \cite{AlexNet}. The reasons behind that are three folds.
  Firstly, rich features extracted from different levels are beneficial.
  Second,  a large amount of training data are used to optimize the tremendous parameters, which is helpful to enhance learning ability of the network. Third,
   very deep neural networks are designed to enhance the fitting ability for the classification task. Roughly, deep learning based approaches can be cast into four categories according to the network modalities, which are 2D-CNNs, Recurrent Neural Network (RNNs), 2D-3D mixed models and 3D-CNNs.

2D-CNNs and recurrent neural network (RNNs) \cite{donahue2015long,lev2016rnn,li2018videolstm} are firstly employed in action recognition. Soon they were  replaced by 3D-CNNs.
3D-CNNs can seamlessly extract features in the  spatial-temporal domain. It proves that 3D-CNNs with spatial-temporal kernels perform better than 2D-CNNs with spatial kernels for action recognition \cite{hara2018can}. C3D is the first one that uses
 3D network for action recognition. Later other variants emerges, and the overall evolution follow the rules from ``shallow'' 3D-CNNs to ``deep'' 3D-CNNs.

Despite the promising performance, two problems should be  solved  in  state-of-the-art action recognition methods mentioned above.

\begin{figure*}[t]
\setlength{\abovecaptionskip}{0.cm}
\setlength{\belowcaptionskip}{-0.cm}
\centering
\includegraphics[width=6in]{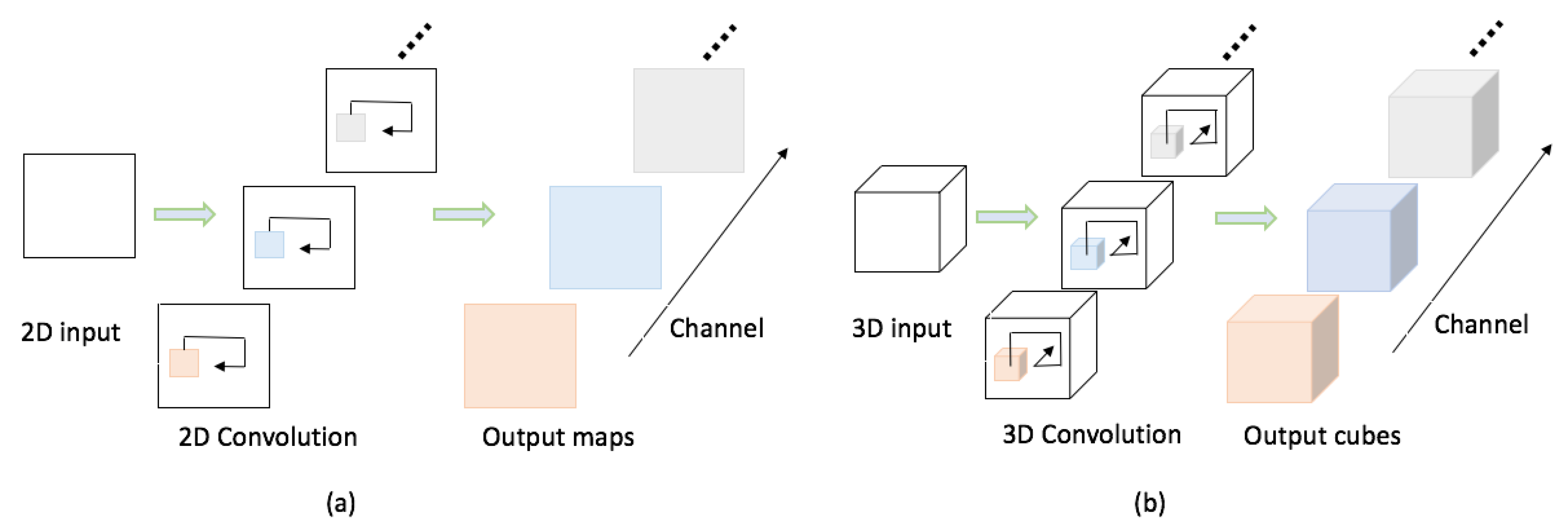}
\DeclareGraphicsExtensions.
\caption{2D and 3D convolution. (a) 2D convolution. During convolution operation, 2D input is convolved with 2D kernels along the 2D spatial dimension; (b) 3D convolution. 3D input is convolved with 3D kernels along the spatial and temporal dimensions. }
\label{fig1}
\end{figure*}

1) \textbf{Depth is beneficial to improve the performance of action classification as compared to using RGB video alone.}
Most existing approaches only take RGB images as inputs, and the appearance and motion cues are usually employed in video sequences to recognize human actions \cite{li2018videolstm, hara2018can, carreira2017quo}. Although impressive advances have been achieved, the lack of 3D structures of the objects as well as the scenes make those approaches struggle to handle the scenarios with  heavy occlusions and similar objects.

With the development of Microsoft Kinect \cite{zhang2012microsoft}, commodity range sensors make it feasible to generate depth at scale.
 Depth provides valuable sources including temporal correlation, emotion expression and motion patterns, which are the key factors for distinguishing human actions under some circumstances. Therefore, depth information can be viewed as a vital complementary to RGB sequence for improving the performance of action recognition.

Some work spent efforts on combining RGB and depth information for action recognition \cite{hu2015jointly, kong2017max, wang2017cooperative}, and demonstrated the effectiveness of modality fusion. Existing networks which use both RGB and depth videos as inputs suffer one of the two drawbacks: the first one is that they usually carry a large amount of parameters and lead to heavy computation. Otherwise with light-weight networks, the performance is  not competitive.

2) \textbf{ Although 3D-CNNs achieved  state-of-the-art performance, they usually contain a large amount of parameters, leading to high computational cost and a demand for large training data.} Among the few existing 3D-CNNs for action recognition, the most representative one is inflated 3D network (I3D) \cite{carreira2017quo}. It is trained on Kinetics \cite{gaidon2013temporal}, and uses a two-stream inflated 3D CNN to achieve the state-of-the-art performance for  action recognition. The success of I3D provides some important hints,  including: First, as the core parts, inception 3D modules provide  possibilities for the network to learn more various and robust features.  Second, with the large-scale training data, deep 3D networks are more likely to achieve performance gain
in terms of classification accuracy.

As a milestone for action recognition, I3D possesses many superior qualities. There still are two disadvantages, which may be further improved.

\textbf{A large amount of  parameters.} Within the I3D network, each inception module is extending 2D-CNN of the inception V1 \cite{ioffe2015batch} to 3D-CNN module.
It contains a large amount parameters and is very computationally heavy. By
 inspecting the network structure, we find that nine inception modules taking up over 80\% of the total parameters, and the computations mainly come from the convolutional operations with 3D kernels. In addition, tremendous parameters in network always accompany very  demanding  memory cost.

Most recently, a trend is to  tailor a large network  into light-weight network. Two representative studies  are ShuffleNet \cite{zhang1707shufflenet} and MobileNet  \cite{howard2017mobilenets}. However, this line of work mainly focuses on 2D-CNNs. It is worth mentioning that depth-wise convolution and group convolution play an important role in designing light-weight networks.

In fact, 3D-CNNs are of  much stronger demand for light-weight design per the aforementioned reasons, yet little work exists for tackling this problem.  3D-CNNs are more complex compared to  2D-CNNs. So are the light-weight operations in 3D-CNNs than that in 2D-CNNs.
Specifically, under the conditions that the depth and width are similar, 3D-CNNs are of higher computational cost and need larger memory storage than 2D-CNNs. Fig.~\ref{fig1} shows the comparison between the 2D convolution and the 3D counterpart. It can be found that their main differences include input dimensions, kernel dimensions and the ways of convolution. For example, for the kernel of size 3, a 2D kernel only has 9 parameters, whereas a 3D kernel has 27 parameters. When convolving a input of size $S$ with a kernel of size 3, the FLOPs of 2D convolution and 3D convolution are ${S}^{2}\times{3}^{2}$ and ${S}^{3}\times{3}^{3}$, respectively. The computational costs of 3D-CNNs are at least one magnitude higher than that of 2D-CNNs, that is $O({n}^{3})$  vs.\  $O({n}^{2})$.

\textbf{Data hungry.} In addition, due to its heavy structure,
 to achieve a better performance,  enormous samples are required during training.
 Although the deeper 3D-CNNs generally have  larger capability,
 3D-CNNs may suffer from over-fitting, if there are not sufficiently large training datasets.

For deep learning based methods,
although using both RGB and depth data generally achieves  better performance than using RGB data alone in action recognition \cite{kong2017max, wang2017cooperative}, most existing RGB-D datasets contain less than 10k videos. The magnitude is far less than that of RGB datasets, such as ImageNet \cite{AlexNet} and Kinetics \cite{kay2017kinetics}. As a result, the lack of training data would inevitably limit the potential of deep neural network.

To summarize,  despite the great efforts and the rapid progress in the past few years,
action recognition is still a challenging task. There is
 much room for improvement. First, we can enrich the input data by employing multimodal data, such as RGB and depth videos. Second, we can optimize the structures of conventional 3D-CNNs by adopting light-weight design.

In this paper, in order to overcome the aforementioned problems in 3D-CNNs, and being inspired by the design of I3D and light-weight 2D-CNNs, we propose a set of light-weight 3D-CNN networks for action recognition using RGB-D data as inputs.  The proposed networks include an inception with spatial and temporal convolution network (IST), a shuffle spatial and temporal convolution network (SST) and a group shuffle spatial and temporal convolution network (GSST), which optimize the 3D-CNNs at
the 3D structure level as well as the channel level.

Our contributions as follows.

\begin{itemize}
    \item In order to reduce the parameter and computation complexity  of 3D-CNNs of  state-of-the-art action recognition framework, we propose a series of light-weight models for action recognition on RGB-D, where RGB information and depth information are jointly used.
    \item We evaluate the proposed light-weight models on two RGB-D datasets, which are the largest RGB-D human activity dataset (NTU RGB+D dataset, NTU) and a small dataset (Northwestern-UCLA Multi-view Action 3D dataset, N-UCLA). Compared to I3D, the proposed models contain much fewer parameters, and are of much lower computational cost.
     More importantly, the proposed models achieve {\em on par}  or
      even better accuracy on both datasets.

\end{itemize}

\section{Related work}
\label{relatedwork}

\subsection{Action recognition}

Action recognition is one of the fundamental problems in human action analysis and receives consistent attention in the computer vision community. Since action cues hide in the spatial-temporal domain, action recognition is typically composed of two ingredients. One is to extract the proper features of the appearance, and the other is to model the dynamic motions \cite{wang2012mining}.

Conventional methods usually based on the handcrafted features such as HOG \cite{laptev2008learning} and HOF \cite{dalal2005histograms,scovanner20073,wang2011action,wang2013action}. Among the  conventional methods, improved dense trajectories (iDT) \cite{wang2013action} achieves the best performance.  Since AlexNet \cite{AlexNet} was invented,
 deep learning has dominated many fields of computer vision, and action recognition is one of them.
 Deep learning based methods for action recognition experienced from 2D-CNNs to 3D-CNNs, and from shallow 3D-CNNs to deep 3D-CNNs.

2D-CNNs \cite{simonyan2014two,wang2015action,feichtenhofer2016convolutional} usually employ a model that is
pretrained on ImageNet \cite{deng2009imagenet} as the initialization. Random sampling or key-frames sampling policies are used
 to extract partial frames from the whole video sequences. To obtain the final classification score, the scores of the sampling frames are often averaged. For shallow 3D-CNNs, studies of \cite{baccouche2011sequential,dalal2006human} are the pioneers that propose to leverage the 3D spatial-temporal kernels in video action recognition. The  Kinetics dataset \cite{kay2017kinetics} makes it possible to train deep 3D-CNNs on large-scale datasets.
Intertwined 3D-2D networks \cite{tran2018closer,zhou2018mict} are shown up as a
 hybrid between 2D-CNNs and 3D-CNNs. It uses 3D spatial-temporal kernels to extract motion details and temporal information
  in the low-level layers and employs  the 2D kernels  to conduct classification and action recognition in the high-level layers.
  A benefit of using such 3D-2D networks, compared to the 3D-CNNs,
  is that the parameters involved in the networks are much reduced. Similarly, pseudo-3D \cite{qiu2017learning} employs
   three parallel 2D residual blocks to extract the spatial and temporal information individually.

\subsection{RGB based action recognition}

Static RGB images in sequential or stacked frames are fed into a neural network to extract appearance features and motion patterns, and 2D-CNNs or RNNs are used for action recognition.
To further improve the accuracy of action recognition, Simonyan and Zisserman proposed to capture the complementary information by incorporating RGB and optical flow information in \cite{simonyan2014two}.
Later,  some methods are  proposed to use RGB and stacked optical flow to form a two-stream network to represent the appearance and motion information \cite{feichtenhofer2016convolutional,wang2015action,wang2015towards,wang2016temporal}. One advantage of these approaches is that 2D-CNNs can be pretrained on ImageNet \cite{deng2009imagenet}.

Recently, a few 3D-CNN structures \cite{hara2018can,carreira2017quo,tran2017convnet} are proposed, trained on large action recognition datasets including Kinetics and Sport-1M. Besides the above work, which focused on ``increasing depth", some models pay
  attention to improve computational efficiency. In the work of ECO \cite{zolfaghari2018eco}, Zolfaghari \emph{et al.} proposed a sampling strategy to sample a fixed number of key frames from videos of different lengths. Once the frames are extracted, they are fed into a 2D-CNN to extract 2D features which are stacked into 3D feature cubes. Afterwords, 3D feature cubes are input to a 3D-CNN to extract deep spatiotemporal features and calculate classification scores. In \cite{zhu2016key} Zhu \emph{et al.} proposed a key frame mining strategy, which share the similar idea of ECO. In \cite{wu2018compressed} Wu \emph{et al.} proposed to use the compressed video to train the network for improving computational speed and saving memory storage.

\subsection{Skeleton based action recognition}

Skeleton based approaches  capture the 3D dynamics.
Conventional methods including conditional random field (CRF) \cite{han2010discriminative} and hidden Markov model (HMM) \cite{lv2006recognition} use 3D points to model the motion in the temporal domain.
 Dynamic temporal warping (DTWF) \cite{muller2006motion} tries to model the spatial-temporal point transition based on nearest neighbor.
  It relies heavily on the metric representing similarity between features and can easily cause misalignment for repeated actions.

Recently, some deep learning based approaches are proposed. In
\cite{wang2012mining}, local occupancy pattern (LOP) is applied to associate 3D joints, and Fourier temporal pyramid (FTP) is used
 to model the temporal information.
 Some other work use recurrent neural network (RNN) or long short term memory network (LSTM) to model the spatial-temporal evolution about the joints \cite{liu2016spatio,liu2017skeleton}.
A drawback of skeleton based methods is that they are  often  sensitive to view transfer and pose variation.

\subsection{Depth based action recognition}

Similar to dense optical flow, scene flow \cite{basha2013multi} calculated from depth data provides the dynamic information of the scene and it is less sensitive than optical flow to the cases of fast-moving and heavy occlusions. Specifically,
the work of  \cite{rahmani2014hopc} and \cite{oreifej2013hon4d} employs
 the 4D features, which are histograms of orientated normal vectors (HON4D) or histogram of oriented principal components (HOPC).
  Some other work employs depth motion map (DMM) to conduct action classification \cite{veeriah2015differential,yang2012recognizing,wang2016action,wang2015convnets}. Specifically, the depth in each frame is aggregated into one single image.
  Based on the aggregating depth, HOG or CNN features are extracted for conducting action recognition.

\begin{figure*}[!t]
\setlength{\abovecaptionskip}{0.cm}
\setlength{\belowcaptionskip}{-0.cm}
\centering
\includegraphics[width=7in]{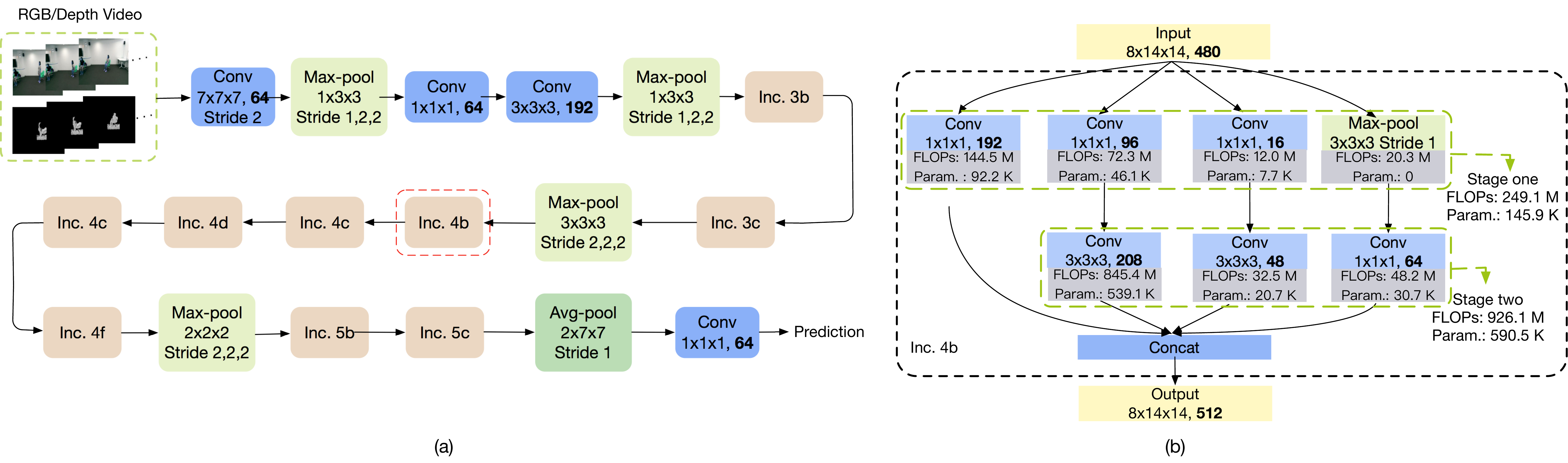}
\DeclareGraphicsExtensions.
\caption{Action recognition with I3D. (a) The overall framework. (b) The structure of the inception module 4b of I3D.
Except the first $7\times 7\times 7$-sized convolution layer, all the strides of the convolution layers used in this structure are set to 1. Batch normalization and ReLu are added to the end of each convolution layer (not shown here).
The last convolution layer  is followed by softmax. In (b), the shape of 3D kernels and the width of each convolution are shown.
 For instance, '$3\times 3\times 3$, \textbf{208}' means that this convolution layer has 208 output channels with $3\times 3\times 3$-sized 3D kernels. The FLOPs of this module are listed on the right side.}
\label{fig2}
\end{figure*}

\subsection{RGBD-based action recognition}

With the development of Microsoft Kinect, cost-effective depth cameras make it possible to capture depth, which opens  new possibilities  for  action recognition.
 Meanwhile, they also introduce some new challenges.
 For example, the acquired depth is usually noisy and it is hard to track the points which fall into the occlusion area \cite{shahroudy2016ntu}.

Compared to RGB videos, RGB-D videos contain the structures of the 3D objects and scenes \cite{han2013enhanced}.
  Asadi-Aghbolaghi \emph{et al.} \cite{asadi2017action} proposed   the method MMDT   to fuse handcrafted features and deep  features for action recognition.
  Specifically, the dense trajectories acquired by iDT and histograms of normal vector (HON) from depth images are extracted to represent the motion patterns. The features extracted from the 2D-CNN are used to represent  the appearance characteristics.
   Wang \emph{et al.} proposed   c-ConvNet \cite{wang2017cooperative}  to employ a joint loss composed of softmax loss and ranking loss to learn homogeneous and heterogeneous features in RGB and depth.

\subsection{Light-weight 2D-CNNs}

For 2D-CNNs, several light-weight networks have been proposed, such as MobileNet \cite{howard2017mobilenets,sandler2018inverted}, ShuffleNet \cite{zhang1707shufflenet}.

In MobileNet, depth-wise separable convolutions are employed to reduce the computational cost in the network, and pointwise convolutions are applied to combine the features of separable channels. MobileNet V2 inherits the advantage of MobileNets and introduces inverted residuals and linear bottlenecks for further improvement.

Taking the structure of bottleneck \cite{he2016deep} as basis, ShuffleNet uses group point-wise convolution and depth-wise convolution to reduce parameters and computational cost. Furthermore, shuffle layer is exploited for information exchange among different channel groups in each shuffle unit.

The above models  heavily relies  on depth-wise convolution and group convolution.

\textbf{Depth-wise convolution.}
 The regular 2D convolution always performs
 convolutions over multiple input channels, and the filter has the same depth with the input. In contrast,
 in the depth-wise convolution, each channel is separate, hence the name ``depth-wise''.
 The three components  are: 1) Split the input as well as the filter into channels. 2) Convolve the input and filter in corresponding channels. 3) Concatenate the output tensors.
 Depth-wise convolution can be viewed as a special form of the inception module. %

\textbf{Group convolution.}
Through sharing parameters among different filter groups, group convolution effectively reduces the redundant parameters.
It can be understood as that group convolution reduces
 parameters by splitting a single convolution layer into several independent sub-paths.

\section{Our method}
\label{methodology}

To achieve light-weight 3D-CNNs for action recognition, we propose three light-weight models including inception with spatial and temporal convolution network (IST), shuffle spatial and temporal convolution network (SST) and group shuffle spatial-temporal convolution network (GSST). Compared to the state-of-the-art method I3D \cite{carreira2017quo},
our three proposed models contain fewer parameters,
 are of lower computation cost and achieves better performance.

Inspired by I3D \cite{carreira2017quo}, the proposed models mainly focus on two aspects of optimization,
which make them the better light-weight 3D networks for action recognition.

\begin{enumerate}[]
\item \textbf{Optimization of the 3D structure}. As mentioned in Section~\ref{introduction}, the additional dimension of 3D-CNNs dramatically increases the computation cost. The goal is to reduce the heavy computation burden due to  the temporal dimension, while still keeping the advantage of 3D-CNNs that operates along both spatial and temporal dimensions.
 One effective approach is to replace the conventional 3D spatio-temporal convolutions with 2D spatial convolutions and 1D temporal convolutions.

\item \textbf{Optimization of the channel dimension}. Motivated by the light-weight  2D-CNNs \cite{zhang1707shufflenet,howard2017mobilenets}, channel shuffling and group convolution are employed to further compress the 3D model along the channel dimension.

\end{enumerate}

In the following, the design principles and details of the proposed light-weight operations are  explained.

\subsection{Separation of 3D convolution into spatial and temporal convolutions}

\begin{figure}[h]
\setlength{\abovecaptionskip}{0.cm}
\setlength{\belowcaptionskip}{-0.cm}
\centering
\includegraphics[width=3.5in]{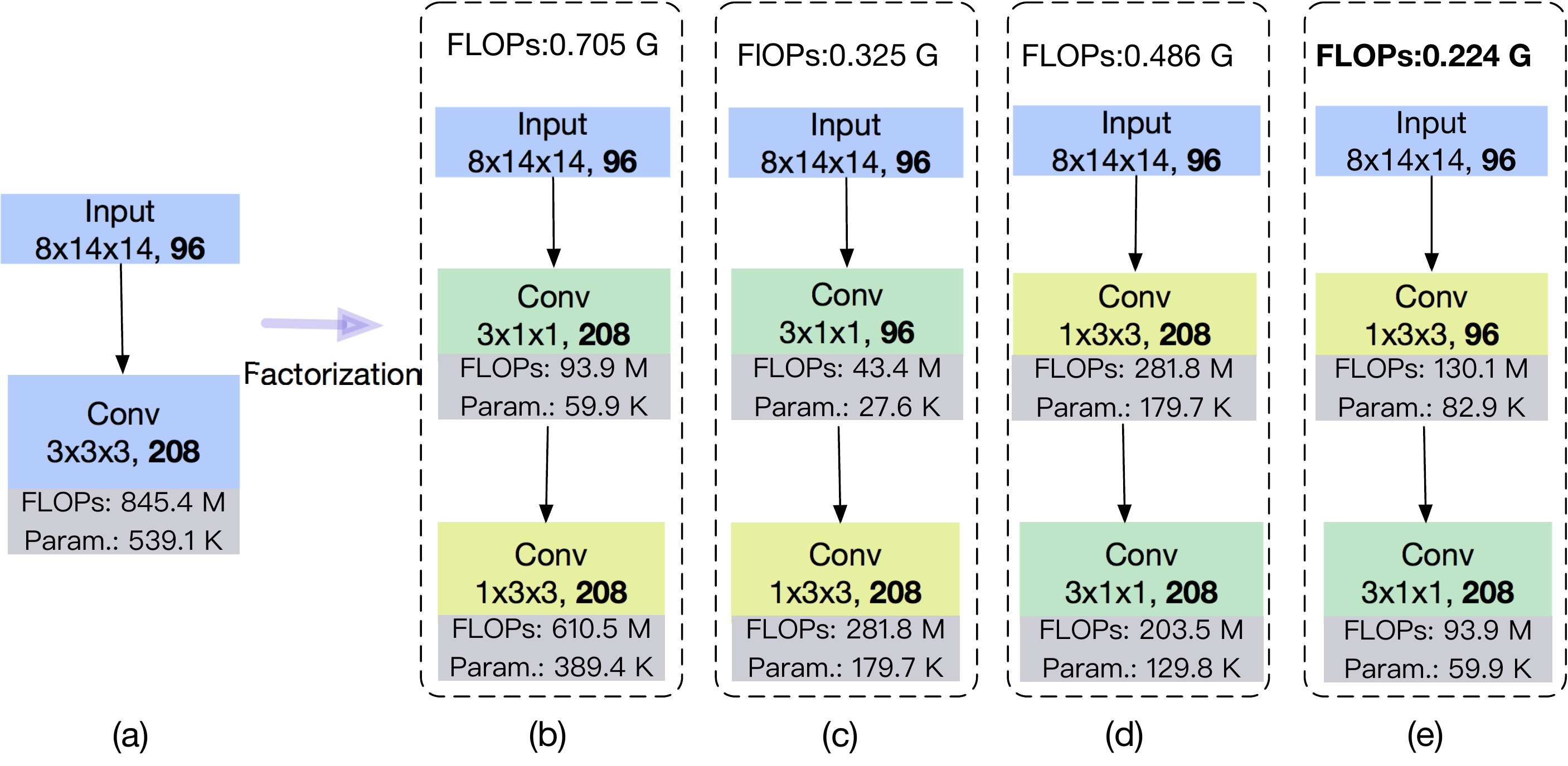}
\DeclareGraphicsExtensions.
\caption{Factorizing 3D convolution into two separate convolutions. (a), 3D convolution, 539136 parameters; (b) factorization structure 1, 449.3k (59.9k + 389.4k) parameters; (c) factorization structure 2, 207.4 (27.6k + 179.7k) parameters; (d) factorization structure 3, 309.5k (179.7k + 129.8k) parameters; (e) factorization structure 4, \textbf{142.8k} (82.9k + 59.9k) parameters.}
\label{fig3}
\end{figure}

\begin{figure*}[t]
\setlength{\abovecaptionskip}{0.cm}
\setlength{\belowcaptionskip}{-0.cm}
\centering
\includegraphics[width=7.2in]{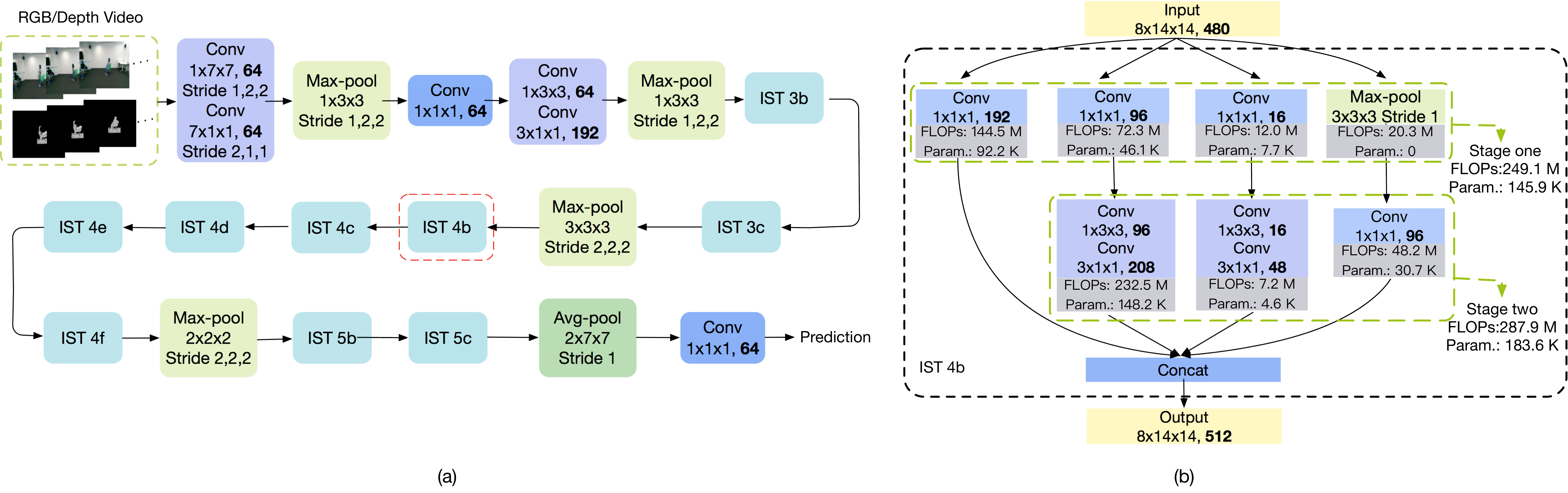}
\DeclareGraphicsExtensions.
\caption{Action recognition with IST. (a) The structure of IST. (b) The detailed structure of IST 4b. In IST, separate convolutions are used in the first three convolution layers and all the Inc.\  modules are replaced with IST modules. In IST 4b, the $3\times3\times3$-sized 3D convolutions are replaced with spatial convolutions of size $1\times3\times3$ and temporal convolutions of size $3\times1\times1$. In order to minimize the number of parameters, the spatial convolution layers keep the same widths as previous layers. For example, from left to right, in the second sub-path, both the widths of first point-wise convolution layer and spatial convolution layer are 96.}
\label{fig4}
\end{figure*}

For ease of exposition, the overall workflow of I3D \cite{carreira2017quo}
 and the structure of its submodule 4b are shown in Fig.~\ref{fig2}.
 It can be seen that the submodule 4b of I3D consists of four sub-paths, which are one single-layer path and three two-layer paths. In order to elucidate easily, the module is divided into two stages, according to the depth of the four subpaths.

 In stage one, the input data is input into three point-wise convolution layers and one max-pooling layer. In stage two, the outputs of two point-wise convolution layers and the max-pooling layer are input into the subsequent respective convolution layers. It can be seen from Fig.~\ref{fig2}(b), the number of FLOPs of stage two is 926.1M, which is about 4 times that in stage one, and takes up 80.19 \% of the total FLOPs in submodule 4b.

Compared to the convolution layers in stage one, those convolution layers in stage two differ as follows,  which increases  the parameters significantly, from 145.9k to 590.5k.

1) The convolution layers in stage two are 3D convolution layers. With the same width, the number of parameters of a $3\times3\times3$-sized 3D convolution layer are 27 times that in point-wise convolution layer. Note that, in both I3D and the proposed networks, biases are omitted, so the parameters of biases are not taken into account here.

2) The width of convolution layers is increased. Specifically, from left to right, the widths of the second path and third path are increased from 96 and 16 to 208 and 48, respectively.

Based on these obversations,
 we  propose two strategies for reducing the parameters.

\begin{enumerate}[]
\item Decomposition of 3D convolution into sequential spatial and temporal convolution. In \cite{szegedy2016rethinking}, Szegedy \emph{et al.} factorized a $n\times n$-sized convolution layer into a $n\times 1$-sized convolution layer followed by a $1\times n$-sized convolution layer and achieved better results.

Thus the number of parameters is reduced while the receptive field keeps unchanged.
Inspired by this operation, we factorize a 3D convolution into two asymmetric convolutions. According to the dimensional order of factorization, the two alternative structures are spatial-temporal and temporal-spatial convolutions. In this paper, we adopt the former because it can compress the parameters to the maximum.

\item Increasing the width of the temporal-convolution. In the module 4b of I3D, the width is increased in stage two. For our design, after the 3D convolution is factorized into the spatial and temporal convolutions, the width can be increased in either the spatial-convolution or the temporal-convolution. We adopt the latter for the purpose of maximal parameter compression. In Fig.~\ref{fig3}, four factorization structures are listed relating to different factorization involving different dimensions, and different locations for increasing the width. In this paper, we employ the factorization structure (e) in Fig~\ref{fig3}, due to the fact that it has the fewest parameters and FLOPs among the four factorization candidates.
\end{enumerate}

Replacing all the 3D convolution layers in the nine modules of I3D, with our new structure, the 3D inception module with spatial and temporal convolution (IST) is obtained. The module 4b of IST is shown in Fig.~\ref{fig4}, where the module 4b of IST is represented as IST 4b.

Compared to I3D, the proposed IST enjoys two advantages.
First, IST contains  fewer  parameters and is of lower computational cost. Specifically, the nine inception modules in I3D take up 80.19 percentage of parameters. From Inc.\ 4b to IST 4b, the number of parameters in the module decrease from 736512 to 329496.
 With the same input, the new structure also decreases the computation cost (FLOPs) by half. Second, IST is more powerful in nonlinearity representation due to the increase in depth. Specifically, from Inc. 4b to IST 4b, the deepest sub-path of convolution layers increases from two to three.

Note that, the very recent method of S3D \cite{xie2018rethinking} which is concurrent to this work,
shows some similarity to the design of the IST module propsed here.

\subsection{Channel shuffling and splitting}

\begin{figure}[t]
\setlength{\abovecaptionskip}{0.cm}
\setlength{\belowcaptionskip}{-0.cm}
\centering
\includegraphics[width=3.2in]{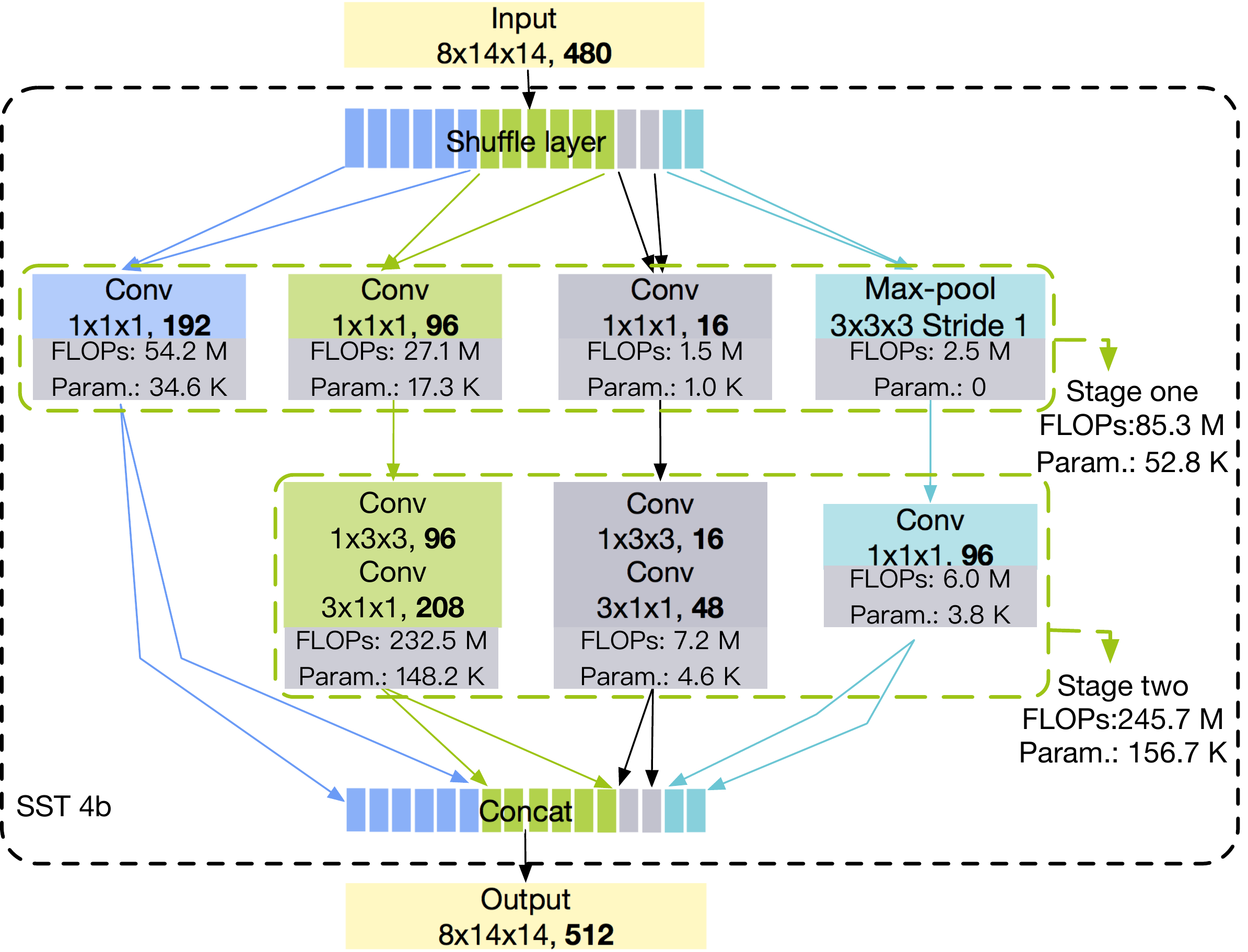}
\DeclareGraphicsExtensions.
\caption{Detailed structure of SST 4b. In the shuffle layer, the input is equally split into 16 groups. Each group has width of 30, then the information of the 16 groups are mixed up by shuffling. According to the capacities of the four sub-paths, the 16 groups are divided into corresponding 4 parts. Each sub-path take one part as input. Compared to IST 4b, the number of parameters of stage one in SST 4b is reduced from 145.9 K to 52.8 K.}
\label{fig5}
\end{figure}

Clearly, IST has less parameters and lower computational cost when compared with I3D. However, the parameters in IST can be further reduced. Our observation is based on the fact that the stage one of each module in IST contains numerous parameters that have not changed during the evolution from I3D. Taking Inc.\ 4b and IST 4b as an example, from Inc.\ 4b to IST 4b, the number of parameters of stage two decreases from 590.5k to 183.6k, while the number of parameters of stage one remains the same, that is 145.9k.

\begin{figure*}[h]
\setlength{\abovecaptionskip}{0.cm}
\setlength{\belowcaptionskip}{-0.cm}
\centering
\includegraphics[width=7in]{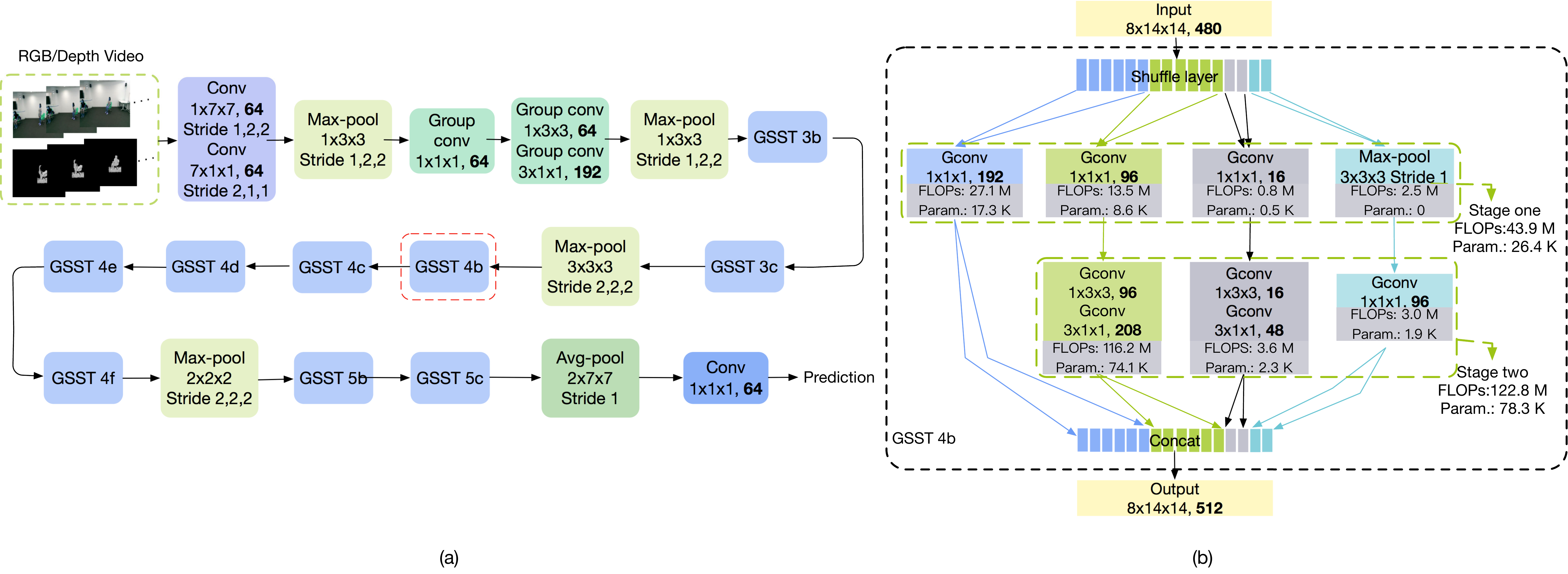}
\DeclareGraphicsExtensions.
\caption{Action recognition with GSST. (a) The structure of GSST. (b) The detailed structure of GSST 4b. In GSST, group convolutions are used in the second and the third convolution layers in the start and all the modules are replaced with the proposed GSST modules.
In GSST 4b, all the convolution layers employed in sub-paths are replaced with group convolution layers.
We set the number of groups in group convolution to 2 in all the GSST modules.
Compared to SST 4b, the number of parameters of GSST 4b is halved.}
\label{fig6}
\end{figure*}

Our work is inspired by ShuffleNet \cite{zhang1707shufflenet}, which is composed of a series of shuffle units.
For each shuffle unit, the point-wise convolution layers are replaced with group point-wise convolution layers for reducing parameters and the shuffle layer is adopted for information flow between different groups. At the beginning of each module, we add a shuffle layer, where the input data is shuffled and then split into four groups based on the capacity of each sub-path. By adding shuffle layer to IST, we obtain our new structure, shuffling spatial and temporal convolution (SST). The architecture of SST 4b is shown in Fig~\ref{fig5}.

\begin{enumerate}[]

\item \textbf{Channel shuffling}. Using channel splitting alone in each module would split the whole network into four independent sub-paths, which may block the information flow between different sub-paths and lead to suboptimal performance.
 In order to overcome this drawback, channel shuffling is used  for information exchange between sub-paths.

\item \textbf{Channel splitting}. In IST modules, there are no point-wise convolution layers, therefore introducing a group convolution layer does not make sense. However, the advantage of the group convolution layer is the  channel-wise sparse connections. In a group convolution, each convolution operates only on the corresponding input channel group rather than the whole set of input channels. In IST, each module has four sub-paths, and each of them operates on all the input channels, which results in dense connections. In contrast,
 in SST, the input data is split into four groups for sparse connection between the input data and the four sub-paths.

\end{enumerate}

Based on the two points described above, the implementation details are as follows.

First, the input is split into $N$ groups by the shuffle layer, and those groups are shuffled  with each other while maintaining the same amount of information with that of before shuffling. The shuffle layer used here follows the one used in \cite{zhang1707shufflenet}, except that the input data is a 3D cube in this paper.

Second, the $N$ groups are divided into four parts corresponding to four sub-paths based on their capacities. Note that, in the I3D and IST modules, the four sub-paths are different in structure and capacity.

Therefore, associating each sub-path with equal amount of input information is not reasonable. The information contained in each sub-path should have a positive correlation with their capacity. Specifically, the number of output channels of the last layer in each sub-path is taken as its approximate estimation of capacity. Suppose that the input is split into $N$ groups and the output channels of the four sub-paths are $C_{1}, C_{2}, C_{3}, C_{4}$. The number of groups that go through sub-path $i$ can be calculated via the formula $N\times{{C}_{i}}/{\sum_{j=1}^{4}{{C}_{j}}}$. We can then fine tune these numbers to ensure that all of them are integers, and that the sum of these numbers should be equal to the number of channels of the input.

After using channel splitting and shuffling, the parameters are decreased from 145.9k in IST 4b  to 52.8k in SST 4b, nearly a third of the original amount.

\subsection{Group convolution}

In related work of light-weight architecture, such as Xception \cite{chollet2017xception}, MobileNet \cite{howard2017mobilenets}, MobileNetV2 \cite{sandler2018inverted}, and ShuffleNet \cite{zhang1707shufflenet},  group convolution and depth-wise convolution play an important role.

In this paper, with the similar purpose of reducing the parameters and computation cost in 3D-CNNs for action recognition,
the structure of SST is further optimized by employing group convolution, which lead our final model, group shuffling spatial and temporal convolution (GSST) network. The workflow of GSST and the architecture of GSST 4b module are shown in Fig~\ref{fig6}.

In Fig~\ref{fig6}(a), GSST replaces all the nine modules of SST with GSST modules.  The second and the third convolutions in each module are converted into group convolutions at the beginning. The group number is set to 2 for all the group convolution layers, and the number of parameters of GSST is reduced by 50 percent compared to that of SST.

\subsection{Light-weight operations summary}

\newcolumntype{C}[1]{>{\centering\let\newline\\\arraybackslash\hspace{0pt}}m{#1}}
\begin{table}[!h]
\setlength{\abovecaptionskip}{0.cm}
\setlength{\belowcaptionskip}{-0.cm}
\renewcommand{\arraystretch}{1.3}
\caption{Light-weight operations}
\label{table1}
\centering
\begin{tabular}{C{1.8cm} C{1.1cm} C{1.1cm} C{1.1cm} C{1.1cm} C{1.1cm}}
\hline
Modules                     & Inc. 4b &   IST 4b     &   SST 4b     &    GSST 4b   \\
\hline
Separating                  &    -    &  \Checkmark  &  \Checkmark  &  \Checkmark  \\
Shuffle layer               &    -    &     -        &  \Checkmark  &  \Checkmark  \\
Group convolution           &    -    &     -        &      -       &  \Checkmark  \\
\hline
Input                       & \multicolumn{4}{c}{$8\times 14 \times 14$, \textbf{480}} \\
Parameter (K)               & 736.5  &   329.5    &    209.5     &    104.7    \\
FLOPS (M)                   & 1175.2 &   537.0    &    331.0     &    166.7    \\
\hline
\end{tabular}
\end{table}

In this section, we briefly summarize the light-weight operations employed in this paper, and analyze the effects  quantitively. Corresponding to Fig~\ref{fig2},~\ref{fig4},~\ref{fig5} and~\ref{fig6}, the information of the parameters and their corresponding light-weight operations used in module 4b are listed in Table~\ref{table1}.

During the evolvement from Inc.\ 4b to GSST 4b, the number of parameters decrease significantly.
First, the number of parameters decrases by 55.26 percent by replacing the 3D convolutions with the separating convolutions in spatial and temporal domains. Second, splitting the input into four parts and adding a shuffle layer further takes away another 36.42 percent of parameters. Finally, converting all the convolutions used in the inception modules into group convolutions further reduces the number of parameters by half.

\subsection{Hyperbolic tangent function based merging strategy}
Since the properties of RGB data and depth data are very different, the recognition performance of the same model with RGB data and depth data also varies a lot. Following \cite{carreira2017quo}, directly averaging the prediction scores of RGB data and depth data does not work when the accuracy of depth data is far away from that of RGB data or \emph{vice versa}. In order to overcome this problem, we propose a robust hyperbolic tangent function based merging strategy.
The hyperbolic tangent function is widely used as an activation function in networks, and we calculate the weights of prediction results via Eq.~\ref{equation1}, where $a$ represents the recognition accuracy on RGB or depth data, and ranges from 0 to 1.

\begin{equation}
\label{equation1}
w = \begin{cases} \frac{{e}^{{a}^{2}}-{e}^{{-a}^{2}}}{{e}^{{a}^{2}}+{e}^{{-a}^{2}}}, \quad a\ge 0.5 \\ \qquad 0, \qquad \rm others \end{cases}
\end{equation}

After the weights are calculated from the hyperbolic tangent function, the perdition scores of RGB data and depth data are merged according to the corresponding weights. The final prediction scores based on the hyperbolic tangent function are denoted as MS 2 in Table~\ref{table2}. The prediction scores obtained  by directly averaging prediction scores of RGB data and depth data are denoted by MS 1 in Table~\ref{table2}.

\section{Experiments}
\label{experiments}
In this section, the proposed models are evaluated and compared with state-of-the-arts on two benchmark datasets, i.e., the NTU RGB+D dataset \cite{shahroudy2016ntu} and the Northwestern-UCLA Multiview Action 3D Dataset \cite{wang2014cross}.

\subsection{Datasets}
The \textbf{NTU RGB+D Dataset (referred to as NTU)} \cite{shahroudy2016ntu} contains more than $56$ thousand video clips of $60$ different types of activities, including daily actions, mutual actions and medical-related actions. This dataset collects activities of $40$ distinct subjects using $3$ Microsoft Kinect v.2 \cite{zhang2012microsoft} cameras concurrently. To our knowledge, NTU is currently the largest RGB-D dataset for action recognition. Following \cite{shahroudy2016ntu}, we perform both the cross-subject and cross-view evaluations. Specifically, for cross-subject evaluation, video clips of the subjects $1$, $2$, $4$, $5$, $8$, $9$, $13$, $14$, $15$, $16$, $17$, $18$, $19$, $25$, $27$, $28$, $31$, $34$, $35$ and $38$ are used for training and the remainder is used for test. As for cross-view evaluation, we used data captured by camera $2$ and $3$ for training and the data from camera $1$ is for test.

N-UCLA \cite{wang2014cross} covers $10$ action categories, and each action is performed by $10$ different subjects. Similar to NTU, N-UCLA is captured by $3$ Kinect v.1 sensors from various viewpoints. Examples from view $1$ and view $2$ are used for training and examples from view $3$ are used for test.

\subsection{Implementation details}

\newcolumntype{C}[1]{>{\centering\let\newline\\\arraybackslash\hspace{0pt}}m{#1}}
\begin{table}[!t]
\setlength{\abovecaptionskip}{0.cm}
\setlength{\belowcaptionskip}{-0.cm}
\renewcommand{\arraystretch}{1.3}
\caption{Results on the NTU RGB+D dataset with Cross-Subject and Cross-View settings. S and D represent skeleton and depth, respectively. (accuracy in \%);( \dag indicates the model is proposed by us).}
\label{table2}
\centering
\begin{tabular}{C{2.5cm} C{1.0cm} C{1.0cm} C{1.0cm} C{1.0cm}}
\hline
Method & Data & CS & CV & Avg\\
\hline
2 Layer P\-LSTM \cite{shahroudy2016ntu} & \multirow{4}*{S} & $62.9$ & $70.3$ & $66.6$ \\
F2CSkeleton \cite{minh2018fine} & & $79.6$ & $84.6$ & $82.1$ \\
ST-GCN \cite{yan2018spatial}  &   & $81.5$  & $88.3$ & $84.9$ \\
SR-TSL \cite{si2018skeleton}   &   & $\textbf{84.8}$  & $\textbf{92.4}$ & $\textbf{88.6}$ \\
\hline
DDMNI \cite{wang2018depth}   & \multirow{6}*{D} & $87.1$ & $84.2$ & $85.7$ \\
Multi-view dynamic images + CNN \cite{xiao2018action} &  & $84.6$ & $87.3$ & $86.0$ \\
I3D      &      & $88.6$ & $91.5$ & $90.1$ \\
IST \dag   &   & $\textbf{89.4}$ & $\textbf{92.5}$ & $\textbf{91.0}$ \\
SST \dag   &   & $88.9$ & $91.3$ & $90.1$ \\
GSST \dag  &   & $88.7$ & $91.0$ & $89.8$ \\
\hline
ResNet50+LSTM \cite{baradel2018glimpse}  & \multirow{6}*{RGB}  & $71.3$ & $80.2$ & $75.8$ \\
Glimpse Clouds \cite{baradel2018glimpse} &   & $86.6$ & $93.2$ & $89.9$ \\
I3D      &   & $89.5$ & $96.6$ & $93.0$ \\
IST \dag   &   & $\textbf{90.8}$ & $\textbf{97.2}$ & $\textbf{94.0}$ \\
SST \dag   &   & $\textbf{90.8}$ & $96.7$ & $93.8$ \\
GSST \dag  &   & $90.6$ & $96.6$ & $93.6$ \\
\hline
\hline
Multiple Stream Networks \cite{garcia2018modality} & \multirow{11}*{RGB+D} & 79.7 & 81.4 & 80.6 \\
DDIf+VDIf+DDIb   & & \multirow{2}*{86.4} & \multirow{2}*{89.1} & \multirow{2}*{87.8} \\
+VDIb(c-ConvNet) \cite{wang2017cooperative} & &       &       &       \\
MS 1, I3D      &   & $92.3$ & $97.4$ & $94.9$ \\
MS 1, IST \dag   &   & $93.0$ & $97.6$ & $95.3$ \\
MS 1, SST \dag   &   & $93.2$ & $97.2$ & $95.2$ \\
MS 1, GSST \dag  &   & $93.0$ & $97.0$ & $95.0$ \\
MS 2, I3D        &   & $92.3$ & $97.4$ & $94.9$ \\
MS 2, IST \dag   &   & $93.0$ & $\textbf{97.6}$ & $\textbf{95.3}$ \\
MS 2, SST \dag   &   & $\textbf{93.2}$ & $97.3$ & $95.2$ \\
MS 2, GSST \dag  &   & $93.0$ &  $97.1$ & $95.1$ \\
\hline
\end{tabular}
\end{table}

\newcolumntype{C}[1]{>{\centering\let\newline\\\arraybackslash\hspace{0pt}}m{#1}}
\begin{table}[!h]
\setlength{\abovecaptionskip}{0.cm}
\setlength{\belowcaptionskip}{-0.cm}
\renewcommand{\arraystretch}{1.3}
\caption{Results on the N-UCLA dataset with three different Cross-View settings (accuracy in \%)}
\label{table3}
\centering
\begin{tabular}{C{2.4cm} C{2.0cm} C{2.0cm} }
\hline
Method             & Data                 & Accuracy \\
\hline
Enhanced viz. \cite{liu2017enhanced} & \multirow{2}*{S}     &  $\emph{86.1}$  \\
Ensemble TS-LSTM   &                      &  $\textbf{89.2}$ \\
\hline
DVV \cite{li2012discriminative}     & \multirow{8}*{D}     &  $58.5$  \\
CVP \cite{zhang2013cross}               &                      & $60.6$  \\
AOG \cite{wang2014cross}               &                      & $45.2$  \\
HPM+TM \cite{rahmani20163d}            &                      & $\textbf{91.9}$ \\

I3D                &                      & $77.7$  \\
IST \dag           &                      & $76.0$  \\
SST \dag           &                      & $70.7$  \\
GSST \dag          &                      & $\emph{78.4}$  \\
\hline
nCTE \cite{gupta20143d}              & \multirow{8}*{RGB}   & $68.6$ \\
NKTM \cite{rahmani2015learning}              &                      & $75.8$ \\
Global Model       &                      & $85.6$ \\
Glimpse Clouds     &                      & $90.1$ \\
I3D                &                      & $92.9$ \\
IST \dag           &                      & $\emph{94.2}$ \\
SST \dag           &                      & $\textbf{95.3}$ \\
GSST \dag          &                      & $93.4$ \\
\hline
\hline
MS 1, I3D                & \multirow{8}*{RGB+D} & $91.2$ \\
MS 1, IST \dag           &                      & $93.8$ \\
MS 1, SST \dag           &                      & $\emph{94.9}$ \\
MS 1, GSST \dag          &                      & $94.0$ \\
MS 2, I3D                &                      & $92.1$ \\
MS 2, IST \dag           &                      & $94.4$ \\
MS 2, SST \dag           &                      & $\textbf{95.5}$ \\
MS 2, GSST \dag          &                      & $93.8$ \\
\hline
\end{tabular}
\end{table}

We use the stochastic gradient descent (SGD) \cite{AlexNet} optimizer (momentum is set to $0.9$) to train all the evaluated models.
The initial learning rate is set to $0.1$ and reduced by $10$ once the training loss is saturated.
For data augmentation, we randomly perform horizontal flipping, spatial crop and temporal clip on the original videos during training.
The video frames are resized to $256\times310$ and then cropped to $224\times224$ patches at four corners or the center.
Each video is randomly clipped to $32$ continuous frames.
For a video shorter than $32$ frames, we repeat it until its length reaches $32$ frames.
During inference, each video is also randomly clipped to several $32$-frame sequences and the corresponding prediction scores are averaged.
The mini-batch size is set to $24$ for RGB videos and $48$ for depth videos.
As the size of N-UCLA is relatively small, we use the model trained on NTU (cross-view data) as the initial model, and fine tune it on the N-UCLA dataset.

\newcolumntype{C}[1]{>{\centering\let\newline\\\arraybackslash\hspace{0pt}}m{#1}}
\newcommand{\tabincell}[2]{\begin{tabular}{@{}#1@{}}#2\end{tabular}}
\begin{table*}[!h]
\setlength{\abovecaptionskip}{0.cm}
\setlength{\belowcaptionskip}{-0.cm}
\renewcommand{\arraystretch}{1.3}
\caption{Comparison of the structures of I3D, IST, SST and GSST.}
\label{table_example}
\centering
\begin{tabular}{C{1.1cm} | C{1.7cm} | C{1.1cm} | C{2.6cm} C{2.6cm} C{2.6cm}C{2.7cm} }
\hline
\multirow{2}*{Layers} & \multirow{2}*{Output size} &  Channel  & \multicolumn{4}{c}{Models} \\
                      &                            &  number   &  I3D  & IST  & SST  & GSST  \\
\hline
Input & $32\times224\times224$ &  3   &  -  &  -   &  -   &  -  \\
\hline
Conv1 & $16\times112\times112$ &  64  & \tabincell{c}{Conv, $7\times7\times7$, 64 \\ stride 2, 2, 2} & \tabincell{c}{Conv, $1\times7\times7$, 64 \\ stride 1, 2, 2 \\ Conv, $7\times1\times1$, 64 \\stride 2, 1, 1 } & \tabincell{c}{Conv, $1\times7\times7$, 64 \\ stride 1, 2, 2 \\ Conv, $7\times1\times1$, 64 \\stride 2, 1, 1 } & \tabincell{c}{Conv, $1\times7\times7$, 64 \\ stride 1, 2, 2 \\ Conv, $7\times1\times1$, 64 \\stride 2, 1, 1 } \\
\hline
Max-p  & $16\times56\times56$  &  64   & \tabincell{c}{Pool, $1\times3\times3$ \\ stride 1, 2, 2} & -  & - & - \\
\hline
\multirow{2}*{Conv2}  & \multirow{2}*{$16\times56\times56$}  & \multirow{2}*{64} & \multirow{2}*{Conv, $1\times1\times1$, 64} & \multirow{2}*{-}  &  \multirow{2}*{-} & \multirow{2}*{Gconv, $1\times1\times1$, 64} \\
       &           &                 &                                  &    &                              \\
\hline
Conv3  & $16\times56\times56$ & 192  & \tabincell{c}{C, $[3,3,3]$, 192 \\ stride 1, 1, 1} &  \tabincell{c}{Conv, $1\times3\times3$, 64 \\ Conv, $3\times1\times1$, 192 } & \tabincell{c}{Conv, $1\times3\times3$, 64 \\ Conv, $3\times1\times1$, 192 } & \tabincell{c}{Gconv, $1\times3\times3$, 64 \\ Gconv, $3\times1\times1$, 192 }\\
\hline
Max-p  & $16\times28\times28$ & 192  & \tabincell{c}{P, $[1,3,3]$ \\ stride 1, 2, 2} & - & - & - \\
\hline
Module group 3 & $16\times28\times28$  & 480  & Inc. 3b - 3c  &  IST 3b - 3c  & SST 3b - 3c  &  GSST 3b - 3c \\
\hline
Max-p  & $8\times14\times14$ &  480   & \tabincell{c}{Pool, $3\times3\times3$ \\ stride 2, 2, 2} & - & - & - \\
\hline
Module group 4 & $8\times14\times14$ &  832   & Inc. 4b - 4f  &  IST 4b - 4f & SST 4b - 4f & GSST 4b - 4f \\
\hline
Max-p  & $4\times7\times7$ & 832     & \tabincell{c}{Pool, $2\times2\times2$ \\ stride 2, 2, 2} & \\
\hline
Module group 5 & $4\times7\times7$ & 1024  & Inc. 5b-5c & IST 5b - 5c  &  SST 5b - 5c  &  GSST 5b - 5c \\
\hline
Avg-p  & $2\times1\times1$ & 1024   & \tabincell{c}{Pool, $[2,7,7]$ \\ stride 1, 1, 1} &  -  &  -  &  -  \\
\hline
Classifier & $3\times1\times1$ & 60 &  Conv, $1\times1\times1$, 60  &   -   &  -  &  - \\
\hline
\end{tabular}
\end{table*}

\subsection{Comparison with state-of-the-art methods}

\textbf{Results on NTU}

On this dataset, we compare the proposed approaches with state-of-the-arts on three input-modality settings, namely depth, RGB and RGB+depth. The results are listed in Table~\ref{table2}.

It can be seen from Table~\ref{table2} that the proposed models achieve the best performance for all the three settings. In terms of recognition accuracy, I3D and the three proposed models outperform others by a large margin. With significantly less computations, the accuracies of our proposed models are still slightly better than I3D.

Following \cite{carreira2017quo}, we combine the results on the RGB and depth data by merging the prediction scores.
Taking the performance difference on RGB and depth data into consideration, we merge the prediction scores using two strategies. With strategy one (denoted as MS 1 in Table III), the prediction scores are averaged. With strategy two (denoted as MS 2 in Table III), the prediction scores are merged with different weights calculated via Eq.~\ref{equation1}.

According to the results on RGB+depth data, we can draw two conclusions.

1) The fuse of prediction scores usually improves the performance of the model.
As we mentioned before, RGB data and depth data are different from each other for carrying different modalities of information. Combining the RGB stream and depth stream is beneficial for the performance of action recognition.

2) Strategy two is slightly better than simple average of the two scores.

\textbf{Results on N-UCLA} We initialize the models with the weights pretrained on NTU, and then fine-tune the pretrained models on N-UCLA. Table~\ref{table3} reports  the fine-tuning results. The results on N-UCLA present a similar trend to that on NTU.

1) Experiments on RGB data achieve much better performance than that on depth data. The gaps in accuracy between RGB data based experiments and depth data based experiments are present to various degrees. On NTU, the gaps range from 2.9 to 3.8 \%. But on N-UCLA, the average gap is about 18.3 \%.

Also, NTU covers 60 different actions and N-UCLA only contains 10 different actions. In terms of the diversity in category, conducting action recognition on N-UCLA is supposed to be much easier than on NTU.

2) The three proposed models obtain best performance on RGB+depth data. The improvements  by combining the prediction results on depth data and RGB data on N-UCLA are not significant as compared to that on NTU. This may be  because the recognition performance on the depth data of N-UCLA is much worse than that of NTU. Therefore, the role that depth data prediction results plays in combining RGB and depth data prediction results for recognition is weakened.

 Using the first strategy to fuse results, the recognition accuracy drops for I3D, IST and SST. With the second strategy,
  combining the prediction scores further improves the recognition accuracy by 0.2\% for IST, 0.2 for SST and 0.4 for GSST.

As can be seen in Table~\ref{table3}, on depth data, HPM+TM shows better performance than our proposed models. However, there are two points that are worth mentioning.

1) On N-UCLA, each class  has only about 97 samples. For deep CNN models, it is very easy to be over-fitted. The shallowest model of our proposed models contain over 20 3D convolution layers. However, HPM+TM used a pretrained 2D-CNN model with only 5 convolution layers.

 2) Comparing with HPM+TM, our method is actually simpler in terms if the design.
 All of our models  are trained and inferred in an end-to-end manner. In HPM+TM, besides training a 2D-CNN model on the synthetic depth pose dataset, it also requires pre-processing\cite{motion_capture} and post-processing\cite{wang2014learning}.

\section{Ablation study}
\label{ablation study}

\begin{figure}[h]
\setlength{\abovecaptionskip}{0.cm}
\setlength{\belowcaptionskip}{-0.cm}
\centering
\includegraphics[width=3.6in]{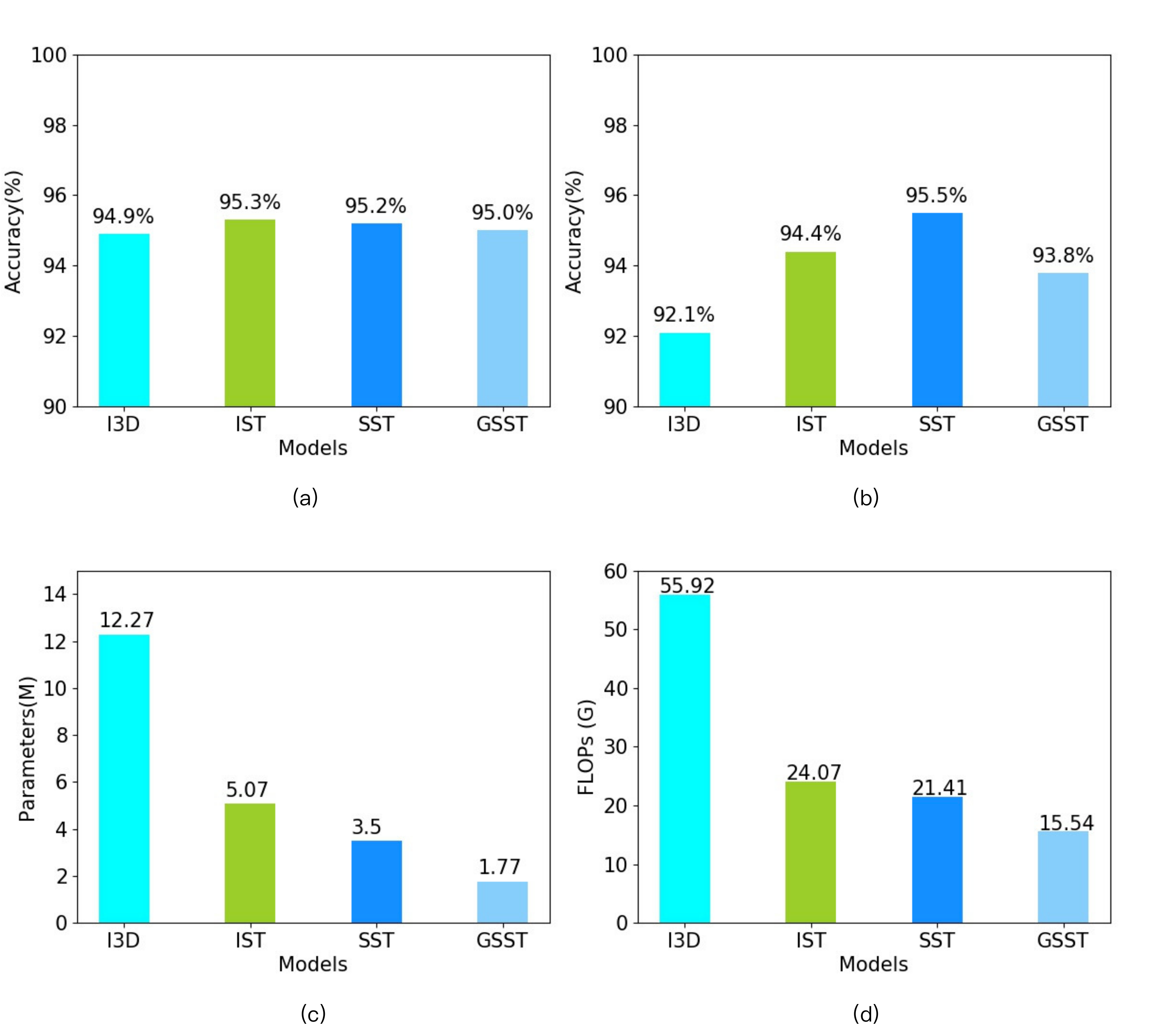}
\DeclareGraphicsExtensions.
\caption{Model comparison in terms of accuracy, number of parameters and computational cost. (a) Performance on NTU; (b) Performance on N-UCLA; (c) Parameters amount; (d) FLOPs}
\label{fig7}
\end{figure}

\newcolumntype{C}[1]{>{\centering\let\newline\\\arraybackslash\hspace{0pt}}m{#1}}
\begin{table*}[!t]
\setlength{\abovecaptionskip}{0.cm}
\setlength{\belowcaptionskip}{-0.cm}
\renewcommand{\arraystretch}{1.3}
\caption{Comparison of the number of parameters and computational cost of I3D, IST, SST and GSST}
\label{table_example}
\centering
\begin{tabular}{C{2.0cm} | C{1.2 cm} C{1.2cm} C{1.2cm} C{1.2cm} | C{1.2 cm} C{1.2cm} C{1.2cm} C{1.2cm} }
\hline
Layers & \multicolumn{4}{c}{Parameters(M)} & \multicolumn{4}{|c}{FLOPs(G)}  \\
\hline
               &   I3D   &  IST   &   SST  & GSST &  I3D  & IST  & SST  & GSST \\
\hline
Input          &  -      & -      &  -     &  -   &  -    &  -   &  -   &  -    \\
\hline
Conv1          &  0.066  & 0.038  &  0.038 &0.038 &13.218 &9.531 &9.531 &9.531  \\

Max-p          &    0    &   0    &   0    &   0  &0.029  &0.029 &0.029 &0.029  \\

Conv2          &  0.004  & 0.004  &  0.004 &0.002 &0.206  &0.206 &0.206 &0.103  \\

Conv3          &  0.332  & 0.074  &  0.074 &0.037 &16.647 &3.699 &3.699 &1.850  \\

Max-p          &    0    &   0    &   0    &  0   &0.022  &0.022 &0.022 &0.022  \\

Module group 3 & 1.222   & 0.493  &  0.402 &0.201 &15.483 &6.340 &5.067 &2.543  \\

Max-p          &    0    &   0    &   0    &  0   &0.020  &0.020 &0.020 &0.020  \\

Module group 4 & 5.894   & 2.354  &  1.647 &0.824 &9.350  &3.799 &2.597 &1.305  \\

Max-p          &    0    &  0     &   0    &  0   &0.001  &0.001 &0.001 &0.001  \\

Module group 5 & 4.754   & 2.103  &  1.333 &0.666 &0.941  &0.421 &0.262 &0.132  \\

Avg-p          &    0    &   0    &   0    &  0   &  -    &  -   &  -   &  -   \\
\hline
Total          & 12.273  & 5.066  &  3.498 &1.768 &55.916 &24.069&21.413&15.536 \\
\hline
\end{tabular}
\end{table*}

\begin{figure*}[h]
\setlength{\abovecaptionskip}{0.cm}
\setlength{\belowcaptionskip}{-0.cm}
\centering
\includegraphics[width=7.2in]{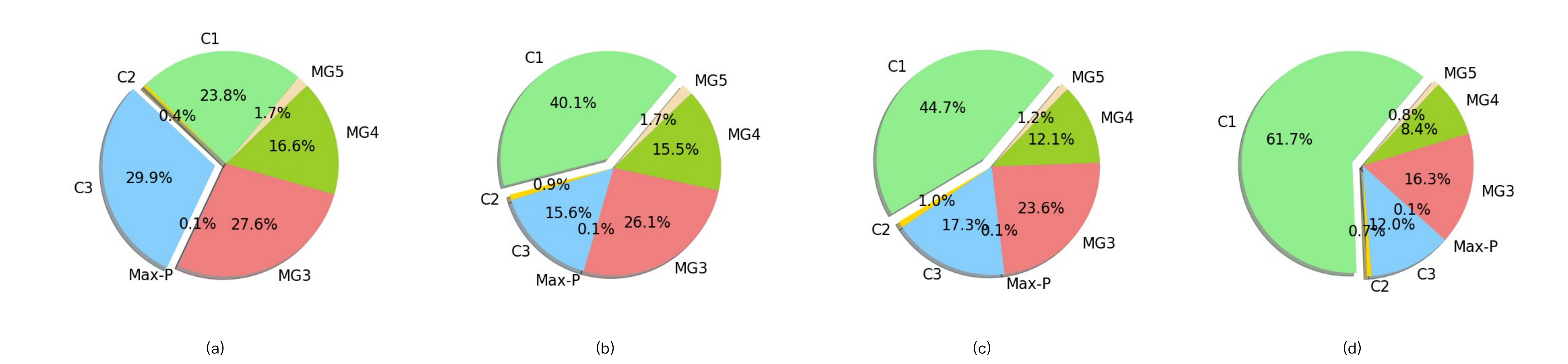}
\DeclareGraphicsExtensions.
\caption{Computation cost decomposition on I3D, IST, SST and GSST. (a) I3D; (b) IST; (c) SST; (d) GSST. For saving space, we represent Conv1, Conv2, Conv3, Module group 3, Module group 4, Module group 5 and the four Map-pooling layer as C1, C2, C3, MG3, MG4, MG5 and Max-P in the figure.}
\label{fig8}
\end{figure*}

\begin{figure*}[h]
\setlength{\abovecaptionskip}{0.cm}
\setlength{\belowcaptionskip}{-0.cm}
\centering
\includegraphics[width=7.2in]{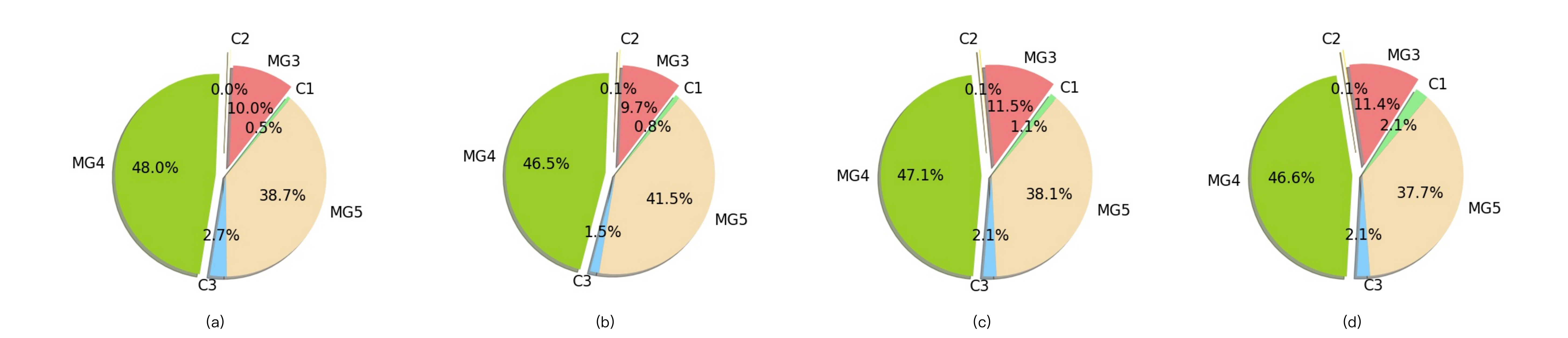}
\DeclareGraphicsExtensions.
\caption{Parameters cost decomposition on I3D, IST, SST and GSST. (a) I3D; (b) IST; (c) SST; (d) GSST.}
\label{fig9}
\end{figure*}

In this section, in order to comprehensively reveal the effects of the compression operations employed in this paper, we conduct a full comparison between I3D, IST, SST and GSST  in terms of
 the action recognition accuracy, network structure, number of parameters, computational cost (FLOPs) and the execution time.

\subsection{Action recognition accuracy}
In the last section, we have compared the four models against  the state-of-the-art. Here, we compare the performance of the four models themselves. In Fig.~\ref{fig7}, the recognition accuracy, parameters and computational cost of the four models are shown.

From Fig.~\ref{fig7}(c) and Fig.~\ref{fig7}(d), we can see that the compression operations that we have designed for compressing I3D are useful. From I3D to GSST, both the number of parameters and computation cost significantly decrease step by step. After a series of compression operations, the number of parameters decreased by a factor of 6.93 times and the computation cost reduced by
a fact of  3.60 times.

Surprisingly, unlike other 2D-CNN compression work, such as MobileNet, and ShuffleNet, where the reduction of network parameters and computation cost usually reduces the classification accuracy, on NTU and N-UCLA, IST, SST and GSST greatly reduce the network parameters and computation cost while improving the acuuracy.

On NTU, the accuracy is increased by 0.4\% in IST, 0.3\% in SST and 0.1\% in GSST. On N-UCLA, the IST, SST and GSST improved the accuracy by 2.3\%, 3.4\% and 1.7\%.

In summary, we do not observe  difference in performance between IST and SST. On NTU, IST has better performance than SST. On N-UCLA, the results are reversed. However, from IST and SST to GSST, the accuracy drops significantly.
 Our conjecture is that this may be because the the employment  of a group convolution may have a negative effect on the capacity of the network and that is why we do not further split each convolution to more groups.

\subsection{Structures}
In Table IV, the detailed structures of I3D, IST, SST and GSST are listed. Gconv represents group convolution. Max-p and Avg-p are the Max pooling layer and average pooling layer, respectively. The size of output data and kernels are represented in the format $t\times h\times w$, where $t$, $h$ and $w$ are the temporal size, height and width of output data or kernels. Conv1, Conv2 and Conv3 represent the first three convoluiton layers in the networks.

For IST, SST and GSST, the parts that have the same structure as the corresponding parts of I3D are shown in '-' for concise display in Table IV.

For the four models, apart from the differences in modules, the rest of the differences are focused on the first three convolution layers. IST, SST and GSST separate Conv 2 and Conv 3 into separate spatial and temporal convolutions. GSST further splits Conv 2 and Conv 3 into group convolutions.

\subsection{The number of parameters and computation cost}
In Table VI, the numbers of parameters and FLOPs in each part of the four models are listed. Fig.~\ref{fig8} and Fig.~\ref{fig9} show the distribution of computation cost and parameters in the first three convolution layers, the three Module groups and the four Max-pooling layers.

As the depth increases, the number of parameters increases and the computation cost decreases. In other words, parameters mainly concentrate on latter layers and computation cost mainly concentrates on the former layers.

From Fig.~\ref{fig8} and Fig.~\ref{fig9}, we can conclude the following two aspects.

 1) As more compression operations are employed, the proportion of computational cost taken by Conv 1 increases from 23.8\% to 61.8\%. Following other light-weight work \cite{zhang1707shufflenet,howard2017mobilenets}, we do not compress the first convolution layer. Therefore, with the decreasing of the total computational cost of the network, the relative computational cost of first convolution layer increases.

  2) The parameters distribution has no much difference between the four models. Note that, the first convolution layer only
  contains 0.5 to 2.1 percent of the total parameters. However, it takes 23.8 to 61.8 percent of total computational cost.

\subsection{Execution time}
In Table VI,  execution time of the four models on different GPU devices are compared. Here, we evaluate the four models on three different GPU devices, Tesla K40c, GeForce GTX TITAN and GeForce GT 740. During inference, as the GeForce GT 740  can take 4 samples at most, we set the batch size to 4 on all the three devices. The numbers listed in TABLE VI are the average time for one iteration. Generally speaking, GSST is twice as fast as I3D in the execution speed.

Note that the improvement in the execution time on  GPUs depends on many factors and  is  less significant than FLOPs.
On other hardwares such as ARM CPUs, the improvement can be more significant.

\newcolumntype{C}[1]{>{\centering\let\newline\\\arraybackslash\hspace{0pt}}m{#1}}
\begin{table}[!t]
\setlength{\abovecaptionskip}{0.cm}
\setlength{\belowcaptionskip}{-0.cm}
\renewcommand{\arraystretch}{1.3}
\caption{Comparison of real-world execution times of the four models on three different GPU devices}
\label{table_example}
\centering
\begin{tabular}{C{2.6cm} C{1cm} C{1cm} C{1cm} C{1cm}}
\hline
\multirow{2}*{Devices}   & \multicolumn{4}{c}{Time cost (s)} \\
                         &   I3D     &   IST    &   SST  &  GSST   \\
\hline
Tesla K40c               &  0.588    &  0.374   &  0.336 &  \textbf{0.296}  \\
GeForce GTX TITAN        &  0.421    &  0.267   &  0.242 &  \textbf{0.214}  \\
GeForce GT 740           &  3.112    &  2.023   &  1.802 &  \textbf{1.618}  \\
\hline
\end{tabular}
\end{table}

\section{Conclusion}
\label{conclusion}
In this paper, we have  introduced 3D-CNNs for  action recognition on RGB-D data  and achieved good performance.
In order to take full advantage of RGB-D data which contains both RGB information and depth information, we take both RGB video and depth video as input into our 3D-CNNs based action recognition frameworks and achieve better performance than only employing single modal video information, i.e. RGB video or depth video alone.

The main contribution of our work is to  overcome the problem that 3D-CNNs generally having a large number of parameters and are of heavy computation. We have proposed  a series of light-weight 3D architectures, namely IST, SST and GSST.

We have validated the proposed models on two RGB-D datasets, the largest available RGB-D human activity dataset, NTU and a small dataset N-UCLA. The proposed light-weight models show comparable recognition accuracy as compared to I3D which is  the state-of-the-art method for  action recognition in accuracy.

Design of  light-weight 3D CNN structures can be very important, not limited to  action recognition.
The designs proposed in this work may be applicable to most 3D data related tasks.

\appendices
\ifCLASSOPTIONcaptionsoff
  \newpage
\fi

\bibliographystyle{IEEEtran}
\bibliography{IEEEabrv,reference}
\end{document}